\documentclass{article}

\usepackage{times}
\usepackage{graphicx} 

\usepackage{natbib}
\usepackage[normalem]{ulem}

\usepackage{algorithm}
\usepackage{algorithmic}

\usepackage{multirow}

\usepackage{hyperref}

\usepackage[accepted]{icml2015} 

\usepackage{mathrsfs}
\usepackage{amsmath,amsfonts}
\DeclareMathOperator*{\argmax}{argmax}
\DeclareMathOperator*{\argmin}{argmin}
\newcommand{\R}{\mathbb{R}}
\newcommand{\E}{\mathbf{E}}

\newcommand{\randn}{\operatorname{randn}}
\newcommand{\orth}{\operatorname{orthogonalize}}
\newcommand{\eig}{\operatorname{eig}}
\newcommand{\mycomment}[1]{}

\icmltitlerunning{Scalable Multilabel Prediction via Randomized Methods}

\begin{document} 

\twocolumn[
\icmltitle{Scalable Multilabel Prediction via Randomized Methods}

\icmlauthor{Nikos Karampatziakis}{nikosk@microsoft.com}
\icmlauthor{Paul Mineiro}{pmineiro@microsoft.com}
\icmladdress{Microsoft CISL,
            1 Microsoft Way, Redmond, WA 98052 USA}

\icmlkeywords{multilabel classification, calibration, embeddings}

\vskip 0.3in
]

\begin{abstract} 
Modeling the dependence between outputs is a fundamental challenge
in multilabel classification.  In this work we show that a generic
regularized nonlinearity mapping independent predictions to joint
predictions is sufficient to achieve state-of-the-art performance
on a variety of benchmark problems.  Crucially, we compute the joint
predictions without ever obtaining any independent predictions, while
incorporating low-rank and smoothness regularization.  We achieve
this by leveraging randomized algorithms for matrix decomposition and
kernel approximation. Furthermore, our techniques are applicable to the
multiclass setting.  We apply our method to a variety of multiclass and
multilabel data sets, obtaining state-of-the-art results.
\end{abstract} 

\section{Introduction}

In multilabel classification the learner is given an example and is
asked to output a subset of the labels that are relevant for that
example. Multilabel problems come up often in web applications where
data of interest may be tagged with zero or more tags representing,
for example, the objects in an image, the topics of a news story,
or the relevant hashtags of a tweet. A simple approach to multilabel
prediction is to learn one classifier per label and predict each label
independently. Why should we expect that we can improve upon this
procedure? One answer is because in many real world multilabel problems
the labels are correlated. A blog post about cooking has a higher chance
of also being related to nutrition than to finance.

In this paper we propose a simple and efficient technique for multilabel
classification, in which the dependencies between predictions are
modeled with a flexible link function.  This corresponds to the rich
literature of using a stacking architecture \cite{wolpert1992stacked}
in the multilabel context to combine independent predictions to
produce a joint prediction \cite{godbole2004discriminative}.  However,
many modern multilabel (and multiclass) problems are characterized by
increasingly large output spaces.  Such large output spaces present both
computational and statistical challenges. The former is easily understood:
a straightforward approach will scale at least linearly with the number
of outputs both during training and inference.   The latter also presents
several challenges:  naive models can have too many parameters, while
certain labels and features can be extremely rare. Predicting each label
must therefore borrow strength from other labels to avoid overfitting.

Our approach to mitigating statistical difficulties is via regularization;
specifically we are interesting in enforcing both low-rank and smoothness.
To mitigate the computational burden of this regularization, we leverage
randomized algorithms for embedding \cite{DBLP:journals/corr/MineiroK14a} 
and kernel approximation \cite{rahimi2007random}.  Fortunately, these procedures can
be combined in a manner which avoids expensive intermediates and scales
to the largest public benchmark data sets.  The technique also applies
to multiclass classification, which is just a special case of multilabel
prediction where exactly one of the labels must be output. To illustrate
this our experiments include both multilabel and multiclass data.

In the rest of the paper we overview a randomized algorithm for embedding
(Section 2), we introduce our technique as fitting a flexible link
function (Section 3), and we argue that statistical (Section 4) and
computational (Section 5) improvements are possible if we use randomized
embedding on the inputs to the link function. We present related work
on Section 6 and experiments on Section 7.

\section{Background and Notation}

Here we will introduce notation and review randomized embedding,
a technique we leverage and adapt in the sequel.  For additional
background on randomized linear algebra, we refer the reader to
\cite{halko2011finding}.

\subsection{Notation}

Vectors are denoted by lowercase letters $x$, $y$ etc.\ and matrices by
uppercase letters $W$, $Z$ etc.  We are given $n$ examples as matrices 
$X \in \R^{n\times d} $ and $Y \in \R^{n\times c}$. We assume that 
the labels $Y$ are sparse: each example is typically associated
with only a handful of labels. We use $||X||_F$ for the Frobenius norm,
$\E[\cdot]$ for the expectation operator.

\subsection{Randomized Embedding}

Algorithm~\ref{alg:rembed} is a recipe for approximating, via randomized
linear algebra techniques, the label projection matrix associated with
the optimal rank-constrained least squares prediction of the labels
given the features \cite{DBLP:journals/corr/MineiroK14a}.  The inputs
are the data matrix $X \in \R^{n \times d}$, the label matrix $Y \in
\R^{n \times c}$, the desired rank $k$, and hyperparameters $\ell$
and $q$.  The algorithm is insensitive to the parameters $\ell$ and $q$
as long as they are large enough (in our experiments we use $\ell=20$
and $q=1$). We start with a set of $k+\ell$ random vectors and use
them to probe the range of $Y^\top \Pi_{X,L} Y$, where $\Pi_{X,L}$
is the projection onto the left singular subspace of the data matrix.
We compute an orthogonal basis for the range and refine it by repeating
$q$ times.   This can also be thought as orthogonal (aka subspace)
iteration for finding eigenvectors with early stopping (i.e., $q$ is
small). Afterwards we use our approximation of the principal subspace
to optimize fully over that subspace (lines 7 and 8) and back out the
solution (line 9).  These last few steps are cheap because we are only
working with a $(k+\ell)\times(k+\ell)$ matrix, and in practice the
most expensive step is the least squares fit to compute $Z$ (line 4).
Note the least squares fit is a regression problem in $k+\ell$ outputs,
independent of $c$, whose computation beneficially exploits the sparsity
of $Y$.

Note once the label projection operator has been obtained, the feature
projection operator can be computed with a final least squares fit using
the projected labels as targets.

\begin{algorithm}[t]
  \caption{Randomized Embedding}
  \begin{algorithmic}[1]
    \STATE {\bf Input:} $X,Y,k,\ell,q$
      \STATE $Q \leftarrow \randn(c, k+\ell)$ 
      \FOR{$i \in \{ 1, \ldots, q \}$} 
        \STATE $Z \leftarrow \argmin_{Z\in \R^{d\times(k+\ell)}} \| YQ - XZ \|^2_F$
        \STATE $Q \leftarrow \orth(Y^\top X Z)$
      \ENDFOR 
      \STATE $F \leftarrow (Y^\top X Q)^\top (Y^\top X Q)$ 
      \STATE $(\tilde{U}, \Sigma^2) \leftarrow \eig(F, k)$
      \STATE $U \leftarrow Q \tilde{U}$ 
      \STATE \textrm{Return} $U$
  \end{algorithmic}
  \label{alg:rembed}
\end{algorithm}

\section{A Flexible Link Function}

We begin with the premise that in many multilabel problems certain label
combinations frequently co-occur. We would like to take advantage of such
co-occurrence patterns, but we cannot jump directly into modeling the
output space because the classifier can only reach that space through
the features it uses. Therefore, we will try to model correlations
among the predictions. The simplest way to do this is to introduce a
link function as an operator that maps a vector of \emph{activations}
$p$ to a vector of predictions $\hat{y}=Cp$. A very popular operator in
the setting of multiclass classification is the softmax: $g(p)=\nabla
\log\sum_i\exp(p_i)$.  This is an oblivious operator whose effect is
to boost the largest activation while ensuring the predictions form a
probability distribution.  In multilabel classification, we would like
to introduce a non-oblivious operator, because we do not know a priori
which labels frequently occur together.

A first possibility is a linear operator i.e.\ $C$ is a $c\times c$
matrix. A natural choice in this case is
\[
C=\frac{1}{n} \sum_{i=1}^n y_i y_i^\top
\]
and with $g_0(p)=Cp$ we see that the link function is taking a weighted
combination of training labels weighted by the inner product between the
activation and each label. Though intuitive, this link function does
not do much: the predictions are a fixed linear transformation of the
activations with no adjustable parameters. This can be easily fixed by
introducing a non-negative scalar $\alpha_i$ for each label:
\begin{equation}
g_1(p)= \sum \alpha_i y_i y_i^\top p,
\label{eq:linkernellink}
\end{equation} so that 
labels with a large $\alpha_i$ influence the prediction more towards
themselves.  This link function is now evidently a linear kernel machine
with $\alpha$ playing the role of dual variables. Our next step is to
generalize the link function by replacing the inner product in input
space with a kernel function, an inner product in a reproducing kernel
Hilbert space:
\begin{equation}
g_2(p)= \sum \alpha_i y_i K(y_i,p)
\label{eq:kernellink}
\end{equation}
In practice we will approximate the kernel function with a finite
sum of basis elements.  There are several techniques for doing this,
including the Nystr\"om method~\cite{williams2001using}, Incomplete
Cholesky Factorization~\cite{fine2002efficient}, and Random Fourier
Features (RFFs)~\cite{rahimi2007random}.  Here we will use RFFs
which are especially easy to work with shift invariant kernels (i.e.\
$K(y,p)=\kappa(y-p)$) and have the advantage that they can be generated in
an oblivious way, without knowing how the activations $p$ look like. This
will prove important for the efficiency of the final algorithm.

Although it is not yet obvious, for our final algorithm we believe
(and present some empirical justification) that shift-invariant kernels
are broadly applicable in this context. Recall that applying $\ell_2$
regularization on a kernel machine with a shift-invariant kernel,
such as the Gaussian kernel, is the same as penalizing the derivatives
of all orders of the functions that the kernel machine can represent
\cite{smola1998learning}. Such a smoothness prior is very appropriate
for our link function: we are applying kernels to the activations (or,
later, to projected activations), not to the feature space.

Recall that for shift invariant kernels
\[
K(y,p)\propto\E_{(r,b)\sim q\times U(0,2\pi)} \cos(r^\top y+b) \cos(r^\top p+b)
\]
where the distribution $q$ depends on the kernel and $U(a,b)$ is the
uniform distribution in $(a,b)$. Therefore we can approximate $g_2$ by
\[
g_3(p)=\sum_{i=1}^n \alpha_i y_i\frac{1}{s}\sum_{j=1}^s \cos(r_j^\top y_i+b_j) \cos(r_j^\top p+b_j)
\]
We now replace $\frac{1}{s}\sum_{i=1}^n \alpha_i y_i \cos(r_j^\top
y_i+b_j)$ by optimization variables $v_j$ and arrive at
\[
g(p)\stackrel{\textrm{def}}{=}\sum_{j=1}^s v_j \cos(r_j^\top p+b_j)
\]
At this point we have the two-stage procedure outlined in
Algorithm~\ref{alg:smooth}. For multilabel problems, the loss function
in the final fit can be either least squares or the sum over classes of
per-class logistic loss.  The latter usually requires a smaller $s$ to
attain the same result, while the former admits a fast fitting procedure
\cite{DBLP:journals/corr/Vincent14} that is independent of the size of
the output as long as the output is sparse. As we argue in the next
sections, Algorithm~\ref{alg:smooth} is actually quite wasteful both
statistically and computationally. In the following sections we will
address these problems using randomized embedding.

\begin{algorithm}[tb]
   \caption{Multilabel via Smooth Link}
   \label{alg:smooth}
\begin{algorithmic}[1]
   \STATE {\bfseries Input:} $x,y$
   \STATE Fit independent binary classifiers: $p_{ij}=f_j(x_i)$
   \STATE Draw random vectors $r_t$ and biases $b_t$ $t=1,\ldots,s$
   \STATE Featurize activations: $\phi_t(x_i)=\cos(r_t^\top f(x_i)+b_t)$
   \STATE Learn weight matrix $V$
   \[
   V=\argmin_{V\in \R^{s\times c}}\textrm{loss}(Y,\Phi(X)V)
   \]
\end{algorithmic}
\end{algorithm}

\section{Dealing with Statistical Issues}

Though we started with assuming that certain combinations of labels
(and activations) are more correlated than others we have not yet
clarified what is the exact property we will be exploiting. Correlations
between activations are captured in their empirical second moment and
our assumption is that this matrix is low rank and therefore can be
described by $k\ll c$ eigenvectors:
\[
\hat{\E}[pp^\top] \stackrel{\textrm{def}}{=} 
\frac{1}{n}\sum_{i=1}^n p_i p_i^\top = \sum_{i=1}^k \lambda_i u_i u_i^\top.
\]
Empirically we have found this assumption to hold for many multilabel
problems. Furthermore, similar ideas, such as assuming that the
second moment of the labels is low rank, have been employed in other
techniques for multilabel problems, e.g., \cite{tai2012multilabel}.
The key benefit of using activations instead of labels is that our
proposed method applies equally well to multiclass problems whereas the
method of \cite{tai2012multilabel} would yield trivial results in the
multiclass setting.

How can the procedure outlined in Algorithm~\ref{alg:smooth} be improved
by taking advantage of the fact that activations are low rank? When
activations are nominally in $\R^c$ but really only span a smaller
subspace of dimension $k$ then the kernel approximation employed
in lines~3 and 4 of Algorithm~\ref{alg:smooth} is very wasteful.
To illustrate this point we will, for simplicity, focus on kernels
that are not only shift invariant, but also rotationally invariant:
$K(p,p')=\kappa(||p-p'||_2)$.  If $p,p'$ only span a space of dimension
$k$ then there exists a $k\times c$ matrix $U^\top$ such that $||p-p'||_2
= ||U^\top p - U^\top p'||_2$. Moreover the matrix $U$ is given by the
top $k$ eigenvectors of $\hat{\E}[pp^\top]$. We can therefore reduce
the dimensionality of the activations $p$ with randomized embedding
before applying our kernel approximation. This reduces the variance, or,
alternatively, requires drawing fewer random vectors to achieve the same
level of approximation.

Since each feature function is now computing $\cos(r_t^\top U^\top p+b)$
the random vectors that we use to project the activations are now $Ur_t$
with $r_t \in \R^k$. Furthermore, their covariance is equal to $U\Sigma
U^\top$ with $\Sigma \in \R^{k\times k}$ being the covariance of the
sampling distribution for $r_t$ (typically a multiple the identity). On
the other hand, Bochner's theorem tells us that there's a one to one
mapping from sampling distributions to positive definite shift invariant
kernels. Therefore, projecting the activations is equivalent to tuning
the kernel to the observed data.

\section{Dealing with Computational Issues}

There are three computational issues with the algorithm as proposed thus
far. First we need to fit individual classifiers for each problem which
can be very time consuming if the number of labels $c$ is large. Second,
when $c$ is large forming the empirical second moment of the activations
and computing the top eigenvectors might be infeasible. Third, the final
optimization over the matrix $V$ still requires the solution of a large
number of problems. Here we address all of them.

We start with the issue of fitting the $s\times c$ matrix $V$.
One possibility is to treat each of the columns of $V$ in parallel as
each of them can be learned independently: by this stage we have finished
modeling dependencies among labels. For square loss we can alternatively
use the recent technique of \cite{DBLP:journals/corr/Vincent14} that shows
how to perform stochastic gradient updates for least squares problems
when the output is large but sparse. This method only requires $O(s^2)$
computation instead of $O(sc)$ where $s$ is the number of basis functions
(i.e., cosines) we use.

We tackle the other two issues together. Our key observation is that in
order to run the algorithm we only need to have the projections of the
activations. Surprisingly, for the case of linear (and kernel) classifiers
it is possible to obtain these without fitting all individual classifiers,
via the randomized embedding procedure previously introduced.

Algorithm~\ref{alg:best} puts the above ingredients together.  Lines 2
through 10 are randomized embedding, which approximates the label
projection operator.  The additional least squares fit of Line 11 yields
the (approximate) feature projection operator.  Lines 12 through 14 are
algorithm \ref{alg:smooth}, but applied in the low-dimensional space.
In particular, each $r_t \in \R^k$.

A few remarks are in order.   For simplicity, we have specified $k$
as a parameter but randomized embedding can also incorporate a strategy
for increasing $k$ if initial estimates of the captured variance are too
low \cite{halko2011finding}. To improve generalization performance, we use a regularized least
squares fit for the randomized embedding solves in lines 5 and 11 of
Algorithm~\ref{alg:best}.  Regularization is less crucial (and sometimes
detrimental) for learning the final $V$ matrix, but this is not surprising
given the regularization inherent in the other aspects of the procedure.

Another (implicit) parameter of the algorithm is the sampling distribution
of the vectors $r_t$. This distribution defines the choice of shift
invariant kernel we will be using to measure similarities between
activations $p$ and labels $y$. Fortunately, we can offer some guidance
here using the spectral properties of various shift-invariant kernels
(see also \cite{le2013fastfood} for details). We recommend using Gaussian
and Cauchy respectively for low and high dimensional $y$. These are
special cases of the multivariate Student distribution with $\nu=\infty$
and $\nu=1$ degrees of freedom.  Intermediate values such as $\nu=3$
and $\nu=5$ can offer better results for medium dimensional $y$.
The corresponding kernels are from the Mat\'ern family.  Some empirical
support for their superiority on medium to high dimensional vectors is
offered in \cite{le2013fastfood}.

\begin{algorithm}[t]
   \caption{Multilabel via Smooth and Low-Rank Link}
   \label{alg:best}
  \begin{algorithmic}[1]
      \STATE {\bfseries Input:} $X,Y,k,\lambda$
      \STATE $(\ell, q) \leftarrow (20, 1)$ 
      \STATE $Q \leftarrow \textrm{randn}(c, k+\ell)$ 
      \FOR{$i \in \{ 1, \ldots, q \}$} 
        \STATE $Z \leftarrow \argmin_{Z\in \R^{d\times(k+\ell)}} \| YQ - XZ \|^2_F + \lambda ||Z||_F^2$ \label{lin:learningstep}
        \STATE $Q \leftarrow \textrm{orthogonalize}(Y^\top X Z)$
      \ENDFOR \label{lin:rangefindend}
      \STATE $F \leftarrow (Y^\top X Q)^\top (Y^\top X Q)$ 
      \STATE $(\tilde{U}, \Sigma^2) \leftarrow \textrm{eig}(F, k)$
      \STATE $U \leftarrow Q \tilde{U}$ 
	\STATE $W=\argmin_{W \in \R^{d\times k}}||YU-XW||_F^2 + \lambda ||W||_F^2$
	\STATE Draw random vectors $r_t$ and biases $b_t$ $t=1,\ldots,s$ 
	\STATE Featurize projected predictions
	\[
	\phi_t(x_i)=\cos(r_t^\top W x_i+b_t)
	\]
	\STATE Learn weight matrix
	\begin{equation}
	V=\argmin_{V\in \R^{s\times c}}\textrm{loss}(Y,\Phi(X)V) + \lambda ||V||^2_F \label{eqn:alg3finalloss}
	\end{equation}
\end{algorithmic}
\end{algorithm}

Algorithm \ref{alg:best} describes training but not inference.
In practice we find Algorithm \ref{alg:best} can be used with multiple
inference procedures if we use a proper scoring rule for equation
\eqref{eqn:alg3finalloss}, such as squared loss\footnote{This is a
poor choice for data sets where some labels are rare.}, logistic loss
(for multiclass), or sum over classes of per-class logistic loss (for
multilabel).  Inference can be per-example for example-averaged metrics
such as Hamming loss or precision-at-$k$, whereas per-class inference
is better for macro-averaged $F_1$ score, a popular evaluation metric in
multilabel tasks.  Inference procedures are discussed in
detail in the experiments.

\section{Related Work}

The algorithms presented here are examples of a stacking
architecture \cite{wolpert1992stacked}, in which the predictions
of (simpler) classifiers are used as input to another model
which produces a (hopefully improved) joint prediction.
In particular, the idea of combining independent predictions
using stacking in the multilabel setting was proposed by
\cite{godbole2004discriminative}, and has been thoroughly explored,
e.g., \cite{montanes2011aggregating,nam2014large,2015arXiv150309022R}.
Our contributions to this line of research are twofold. First, we
empirically demonstrate that kernel machines, with shift invariant kernels
approximated in the primal, provide an effective and computationally
convenient choice for the final classifier. Second, we train our pipeline
without fitting $c$ independent classifiers as a first step.

Our technique can also be viewed as a calibration procedure, as the
second stage is nothing but a link function mapping independent
predictions to joint predictions. Many calibration procedures
\cite{platt1999probabilistic,zadrozny2001obtaining,kakade2011efficient}
have focused on binary classification and they are now widely used in
applications together with diagnostic tools such as calibration plots.
Calibration for multiclass and multilabel classification has received
little empirical attention. A notable exception for multiclass is
\cite{zadrozny2002transforming} which first produces calibrated
probability estimates for induced binary problems and then combines
these estimates to a final estimate of the posterior probability of
each class. However, it is not clear how well calibrated the final
multiclass estimates are.  In high dimensions, given hardness results
\cite{hazan2012weak} and a lack of diagnostic tools, our approach follows
a more pragmatic route: select a family of flexible link functions via
a kernel machine parameterization, then learn an efficient approximation
to a good link function in that family using random features.

The use of embeddings is pervasive in the multilabel
literature, as it provides both computational and statistical
benefits. \cite{weston2002kernel} embed both features and labels
into a common vector space, and use pre-image search for inference.
Our procedure can be considered a more efficient analog of this where the
search is approximated by inference.  \cite{hsu2009multi}, motivated by
advances in compressed sensing, utilized a random embedding of the labels
along with a sparse decoding strategy.  Our embedding is not oblivious
and takes into account both inputs and outputs, and in our experimental
section we demonstrate a resulting improvement.  \cite{bengio2010label}
combined a tree based decomposition with a low-dimensional label
embedding, jointly learning both.  End-to-end fine-tuning of our
architecture is conceptually straightforward and plausibly useful,
but this is outside the scope of this paper.  The principal label space
transform (PLST)~\cite{tai2012multilabel} constructs a low-dimensional
embedding using principal components on the empirical label covariance,
which is then utilized along with a greedy sparse decoding strategy.
Because it uses the empirical label covariance, PLST is not applicable
to the multiclass setting.

The conditional principal label space transformation (CPLST), another
dimensionality reduction approach to multilabel classification, has
strong connections to our technique.  In particular \citep{cpslt2012}
initially SVD the same matrix as in algorithm \ref{alg:best}, denoted
here as $Y^\top X Z^*$, albeit without leveraging randomized techniques.
The similarities are intriguing given that CPLST is motivated by a
bound on Hamming loss.  However, CPLST solves an optimization problem
which is designed to make an independent decode strategy effective, and
apply kernelization to the feature space; whereas we learn a decoder
and apply kernelization to the decoding problem, i.e., predictions.
We attribute the broad applicability of shift-invariant
kernels on the output to the shared statistical structure of
multilabel problems encountered in practice, as opposed to the diverse
statistical structure of features, e.g., sparse high-cardinality text
vs. dense low-cardinality images.  In other words, the choice of kernel
in our procedure is greatly simplified.

Tree-based approaches are another major category of multilabel learning
algorithms.  Due to the richness of the literature, we refer the
reader to a survey paper~\cite{cerri2014extensive}.  Here we discuss
FastXML~\cite{Prabhu14}, a multilabel tree ensemble method for which
we have direct experimental comparisons.  FastXML makes several design
choices to mitigate the computational expense of applying decision
trees to high label cardinality (aka extreme) multilabel problems.
In particular, FastXML partitions the feature space, in contrast to some
approaches that partition the label space.  Furthermore, FastXML also
avoids solving an expensive label assignment problem at each node by
using the union of the labels encountered in the training set (ordered
by empirical frequency).  This yields state of the art performance on
multiple datasets when using a precision-at-$k$ metric.

\section{Experiments}

Unless otherwise indicated, confidence intervals on test set metrics
are 90\% confidence intervals and are estimated using the bootstrap
\cite{efron1994introduction}.

\subsection{Small Benchmark Datasets}

We begin by demonstrating the effectiveness of the technique
on several multilabel data sets from mulan.sourceforge.net
\cite{tsoumakas2010mining}, utilizing their train-test splits.
Table~\ref{tab:data} lists the details of these datasets, which span
several application domains.  All datasets are used as-is without
preprocessing.

\begin{table}[t]
\caption{Dataset details.}
\vskip 1ex
\label{tab:data}
\centering
\begin{small}
\begin{sc}
\begin{tabular}{l|c| c|c|r}
Dataset & Domain & Examples & Inputs & Outputs \\ \hline
\multicolumn{5}{|c|}{mulan.sourceforge.net} \\ \hline  
\abovespace
bibtex     & text & 7395  & 1836 & 159 \\
corel5k    & image & 5000  & 499  & 374 \\
mediamill  & video & 43K   & 120  & 101 \\ 
yeast  & biology & 2417   & 103  & 14 \\ \hline
\multicolumn{5}{|c|}{other} \\ \hline
industries & text & 23K   & 47K  & 354 \\ 
odp & text & 1.5M & 0.5M & 100K \\
lshtc & text & 2.4M & 1.6M & 325K \\ 
\end{tabular}
\end{sc}
\end{small}
\end{table}

We compare our approach with the best reported results from the
multilabel survey paper of \cite{metz2015comparing}, which considered
1543 publications using the mulan datasets, disqualifying all but 64
for reasons such as duplicate papers with different titles, or using
preprocessing which is not publicly available.  We further confirmed that
the best reported result from \cite{metz2015comparing} was at least as
good as the results reported for PLST and CPLST.  The metrics we used
are example-based Hamming Loss (HL) and label-based macro-averaged $F_1$
($F_1^M$), defined in the standard way \cite{metz2015comparing}.

Training was via Algorithm~\ref{alg:best}, with sum over classes of
per-class logistic loss used for equation \eqref{eqn:alg3finalloss}.
We used a random feature approximation to a Gaussian kernel, introducing
two additional hyperparameters (bandwidth and number of basis functions).
Optimization of equation \eqref{eqn:alg3finalloss} was via preconditioned
SGD with momentum, introducing four additional optimization-related
hyperparameters (learning rate, learning rate decay, momentum, and number
of data set passes).  The embedding dimension $k$ was chosen to capture
90\% of the empirical label covariance.  All other hyperparameters were
determined by extracting a 10\% holdout set from the original training set
and optimizing via random search \cite{bergstra2012random}; once hyperparameters were determined
a final assessment was done using the original train-test split.
Note hyperparameters were tuned separately for each metric.

\begin{algorithm}[t]
   \caption{$F_1$ inference for a single class. $\hat Z$ are the predicted label probabilities for the test set, and $p$ is the empirical frequency of the label on the training set.}
   \label{alg:F1infer}
  \begin{algorithmic}[1]
      \STATE {\bfseries Input:} $\hat Z,p$
      \STATE Estimate number of positives: $n_+ \leftarrow p | \hat Z |$
      \STATE Sort probabilities: $\hat Z_{\textrm{sorted}} \leftarrow \mathrm{sort} (\hat Z, \mathrm{'descend'})$
      \STATE Estimate denom: $r \leftarrow \{ n_+ + 1 , \ldots, n_+ + | \hat Z_{\textrm{sorted}} | \}$
      \STATE Estimate f: $\hat f \leftarrow 2\ \mathrm{cumsum} (\hat Z_{\textrm{sorted}}) ./ r$
      \STATE Maximize estimate: $m^* = \argmax \hat f$
      \STATE Threshold at $m^*$: $\hat Y \leftarrow (\hat Z \geq \hat Z_{\textrm{sorted}} (m^*))$
      \STATE {\bfseries Return} $\hat Y$
\end{algorithmic}
\end{algorithm}

\begin{table}[t]
\caption{Mulan Test Set Metrics}
\label{tab:results}
\vskip 1ex
\centering
\begin{small}
\begin{sc}
\begin{tabular}{c|c|c|c}
Metric & Dataset & Survey & This Paper \\ \hline
\multirow{3}{*}{HL $(\downarrow)$} & bibtex & 0.01 & [0.0138, 0.0145] \\
& corel5k & 0.01 & [0.0102, 0.0106] \\
& mediamill & 0.03 & [0.0310,0.0316]  \\
& yeast & 0.30 & \textbf{[ 0.212, 0.226 ]} \\ \hline
\multirow{3}{*}{$F_1^M$ $(\uparrow)$} & bibtex & 0.32 & [0.301, 0.333] \\ 
& corel5k & 0.04 & \textbf{[0.0511, 0.0583]}\\ 
& mediamill & 0.19 & \textbf{[0.238, 0.258]} \\ 
& yeast & 0.87 & [0.428,0.450] \\ \hline
\end{tabular}
\end{sc}
\end{small}
\end{table}

Inference is done per-example for Hamming Loss, and per-class for
$F_1^M$.  Per-example inference merely thresholds the predicted
probability of each class at $1/2$.  Per-class $F_1$ inference is described
in Algorithm~\ref{alg:F1infer}, which is a simplified version of the $F_1$
inference procedure in \cite{dembczynski2013optimizing}.  From Table
\ref{tab:results} we conclude our approach is typically competitive with
the best reported results and sometimes superior.

\subsection{Is it Just About Flexibility?}

Our approach seems to be working well in practice but a reasonable
question at this point is where is the better performance stemming
from? Since we have a two stage procedure with non-linear features at the
second step, could it be the case that we could have just obtained the
same results by a much simpler method? Here we illustrate that simple
methods such as blindly using a kernel (approximation) directly to the
inputs and predicting the labels independently produce very different
results than our judicious use of flexibility to model inter-label
dependencies.

We focus on the RCV1 industries dataset that has a small training set,
compared to its number of features. Based on this, we should
expect that naive application of flexible modeling can lead to decreased
generalization performance. Indeed, we performed three experiments:
learn individual logistic regressions to predict each of the 354
categories, learn individual kernel logistic regressions with a Gaussian
kernel approximation (the bandwidth was selected as in the previous
experiments), and Algorithm \ref{alg:best} with squared loss for equation
\eqref{eqn:alg3finalloss}.  The best test Hamming loss of $0.00159$ was
attained by Algorithm \ref{alg:best}.  Second was the independent logistic
regression with a loss of $0.00163$. Finally, learning independent kernel
logistic regressions had a loss of $0.00193$. Using the bootstrap, we
obtained confidence intervals at most $\pm 7\cdot 10^{-6}$ so none of
the confidence intervals overlap.

\subsection{ODP}

The Open Directory Project \citep{ODP} is a public
human-edited directory of the web which was processed by
\cite{bennett2009refined} into a multiclass data set.  Here we compare
with \cite{choromanska2014logarithmic}, utilizing the same train-test
split, features, and labels.  Specifically there is a fixed train-test
split of 2:1 for all experiments, the representation of document is a
bag of words, and the unique class assignment for each document is the
most specific category associated with the document.

Training was via Algorithm~\ref{alg:best}, with logistic loss used for
equation \eqref{eqn:alg3finalloss}. We used a linear kernel, i.e.,
in lieu of lines 12 and 13 we merely use the embedding directly for
equation \eqref{eqn:alg3finalloss}.  Essentially this is rank-constrained
logistic regression, i.e., we are only exploiting the regularization
of Algorithm~\ref{alg:best} rather than the increased flexibility.
Optimization of equation \eqref{eqn:alg3finalloss} was via preconditioned
SGD with momentum, introducing four additional optimization-related
hyperparameters (learning rate, learning rate decay, momentum, and number
of data set passes).   The embedding dimension and other hyperparameters
were tuned by hand on a 10\% validation set extracted from the training
set.

\begin{table}[t]
\caption{ODP results.  $k=300$ for both embedding strategies.}
\vskip 1ex
\centering
\begin{small}
\begin{sc}
\begin{tabular}{|c|c|c|c|c|} \hline
Method & Alg~\ref{alg:best} & CS + LR & LT \\ \hline \abovespace \belowspace
\parbox[c]{40pt}{\centering Test \\ Error~(\%)} & [83.21, 83.39] & [85.39, 85.56] & 93.46 \\ \hline
\end{tabular}
\end{sc}
\end{small}
\label{tab:odp}
\end{table}

In Table~\ref{tab:odp} we compare against Lomtree (LT), which has
training and test time complexity logarithmic in the number of
classes~\citep{choromanska2014logarithmic}.  LT is provided by the
Vowpal Wabbit~\citep{langford2007} machine learning tool.  We also
compare against a compressed sensing analog of Algorithm~\ref{alg:best}
(CS + LR), where the matrix $U$ computed in lines 2 through 10 is 
replaced by a Gaussian random matrix.  Both variants of Algorithm~\ref{alg:best}
outperform Lomtree, which is unsurprising given Lomtree's emphasis on
computational complexity.  Furthermore, there is a lift from using a
tuned embedding instead of a random one.

To the best of our knowledge, this is the best published result on
this dataset.

On a standard desktop\footnote{dual 3.2Ghz Xeon E5-1650 CPU and 48Gb
of RAM.} it takes approximately 6000 seconds to compute the embedding
(lines 2-11 of Algorithm~\ref{alg:best}), of which 3000 seconds is due
to line 11 (which is also necessary with compressed sensing). Each data
pass for optimizing equation \eqref{eqn:alg3finalloss} takes 1200 seconds,
and validation error starts to turn up after 30 iterations.

\subsection{LSHTC}

The Large Scale Hierarchical Text Classification Challenge (version 4) was
a public competition involving multilabel classification of documents into
approximately 350,000 categories \citep{LSHTC4}. The training examples
and labels and the test examples are available from the Kaggle platform.
The features are bag of words representations of each document.

Training was via Algorithm~\ref{alg:best}, with sum over classes of
per-class logistic loss used for equation \eqref{eqn:alg3finalloss}.
We found Cauchy distributed random vectors, corresponding to the
Laplacian kernel, gave good results.  Optimization of equation
\eqref{eqn:alg3finalloss} was via preconditioned SGD with momentum,
introducing four additional optimization-related hyperparameters (learning
rate, learning rate decay, momentum, and number of data set passes).
The embedding dimension and other hyperparameters were tuned by hand,
although we were ultimately limited by available memory on our standard
desktop\footnotemark[2], i.e., it appears a larger (than 800) embedding
dimension and more (than 4000) basis functions would further improve
results.

We used per-example inference, thresholding each classes predicted
probability at $1/2$.  This produces poor results on macro-averaged F1,
which is what the Kaggle oracle uses to evaluate submissions.  However it
does well on example-based metrics.  We compare with published results
of \cite{Prabhu14}, who report example-averaged precision-at-$k$ on the
label ordering induced for each example.  To facilitate comparison we
do a 75:25 train-test split of the public training set, which is the
same proportions as in their experiments (albeit a different split).

\begin{table}[t]
\centering
\caption{LSHTC results.}
\vskip 1ex
\begin{small}
\begin{sc}
\begin{tabular}{|c|c|c|c|} \hline
Method &  Alg~\ref{alg:best} & FastXML & LPSR-NB \\ \hline
\parbox[c]{80pt}{\centering Test\\Precision-at-1 (\%)}& 53.39 & 49.78 & 27.91 \\ \hline 
\end{tabular}
\label{tab:lshtc}
\end{sc}
\end{small}
\end{table}

Table~\ref{tab:lshtc} contains the results.  LPSR-NB is the Label
Partitioning by Sub-linear Ranking algorithm of \cite{weston2013label}
composed with a Naive Bayes base learner, as reported in \cite{Prabhu14},
where they also introduce and report precision for the multilabel tree
learning algorithm FastXML.

\section{Conclusions}

In this paper we have proposed a procedure for learning a flexible
regularized link function for multilabel (as well as multiclass) problems.
Our procedure empirically works better than many strong baselines and
scales to large output spaces.  Thus, similar to \cite{2015arXiv150309022R},
we conclude that simple label dependency models lead to state of the art
computational and statistical performance.

In the future we plan to investigate the applicability and effectiveness
of analogous procedures in more complex output spaces where the dependency
structure of the output variables is specified by a graphical model.

\section*{Acknowledgments} 
 
We thank John Langford for providing the \mycomment{ALOI and }ODP data set\mycomment{s}.

\bibliography{multilabel}
\bibliographystyle{icml2015}

\end{document}